\documentclass[letter, 10 pt, conference]{ieeeconf}  
\IEEEoverridecommandlockouts                              

\usepackage{graphicx}
\usepackage{latexsym}

\usepackage{amssymb}
\usepackage{amsmath}
\usepackage{bm}
\usepackage[hidelinks]{hyperref}
\usepackage{units}

\usepackage{tikz}
\usepackage{pgfplotstable}
\usepackage{pgfplots}
\pgfplotsset{compat=newest}
\pgfplotsset{minor grid style={dashed,white!80!black}}
\pgfplotsset{major grid style={white!80!black}}
\newcommand\copyrighttext{%
  \footnotesize \textcopyright 2020 IEEE. Personal use of this material is permitted. Permission from IEEE must be obtained for all other uses, in any current or future media, including reprinting/republishing this material for advertising or promotional purposes, creating new collective works, for resale or redistribution to servers or lists, or reuse of any copyrighted component of this work in other works.
  DOI: \href{https://ieeexplore.ieee.org/document/9304733}{10.1109/IV47402.2020.9304733}}
\newcommand\copyrightnotice{%
\begin{tikzpicture}[remember picture,overlay]
\node[anchor=south,yshift=10pt] at (current page.south) {\fbox{\parbox{\dimexpr\textwidth-\fboxsep-\fboxrule\relax}{\copyrighttext}}};
\end{tikzpicture}%
}

\title{\LARGE \bf
An Entropy Based Outlier Score and its Application to Novelty Detection for Road Infrastructure Images}

\author{Jonas Wurst$^{1}$, Alberto Flores Fern\'{a}ndez$^{1}$, Michael Botsch$^{1}$ and Wolfgang Utschick$^{2}$
\thanks{$^{1}$CARISSMA, Technische Hochschule Ingolstadt, 85049 Ingolstadt, Germany
        {\tt\small\{firstname.lastname\}@thi.de}}%
\thanks{$^{2}$Technical University Munich, 80333 Munich, Germany        {\tt\small utschick@tum.de}}%
}

\begin{document}

\maketitle
\copyrightnotice
\thispagestyle{empty}
\pagestyle{empty}

\begin{abstract}
	A novel unsupervised outlier score, which can be embedded into graph based dimensionality reduction techniques, is presented in this work. The score uses the directed nearest neighbor graphs of those techniques. Hence, the same measure of similarity that is used to project the data into lower dimensions, is also utilized to determine the outlier score. The outlier score is realized through a weighted normalized entropy of the similarities. This score is applied to road infrastructure images. The aim is to identify newly observed infrastructures given a pre-collected base dataset. Detecting unknown scenarios is a key for accelerated validation of autonomous vehicles. The results show the high potential of the proposed technique. To validate the generalization capabilities of the outlier score, it is additionally applied to various real world datasets. The overall average performance in identifying outliers using the proposed methods is higher compared to state-of-the-art methods. In order to generate the infrastructure images, an openDRIVE parsing and plotting tool for Matlab is developed as part of this work. This tool and the implementation of the entropy based outlier score in combination with Uniform Manifold Approximation and Projection are made publicly available.
\end{abstract}
\section{INTRODUCTION}
The safety of autonomous cars will be one of the key factors for their success \cite{Fraade-Blanar2018a}. However, to demonstrate with statistical significance, that an autonomous vehicle is performing 20 percent better than the average human driver would require the car to drive more than 5 billion miles \cite{Kalra2016a}. Since this approach is infeasible, other strategies, such as scenario based validation, should be used and combined with other strategies \cite{Junietz2018a}. In the scenario based validation, the capabilities  of an autonomous vehicle are demonstrated using some representative scenarios. They can for example be designed manually, or determined based on simulations. Another possibility is to identify relevant scenarios from real world driving. Besides representative scenarios, novel scenarios should be considered as relevant for the validation purpose.

Identifying novel scenarios can be interpreted as an outlier detection task, where a newly observed scenario is tested with respect to its novelty or outlierness. In this work a novel nearest neighbor graph based outlier detection method is introduced. Hence, this work contributes to the question of how to identify novel scenarios, which is crucial for the validation of autonomous vehicles.

A traffic scenario consists of various components. Besides the dynamic objects, the infrastructure forms an important part. In this work, the outlier method is applied to infrastructure images to show its performance capabilities. As in \cite{Nadarajan2017a} and \cite{Chaulwar2017b}, one can think of stacking various layers of images, containing objects, their dynamics, the infrastructure and further information. Here, just the infrastructure layer is used as a case study.

The novel outlier score is based on a directed nearest neighbor graph. Using graphs allows one to use the data points itself to determine outliers. Such a graph is constructed also in dimensionality reduction techniques as \textit{t-Distributed Stochastic Neighbor Embedding} (t-SNE) \cite{Maaten2008a}, Barnes-Hut-SNE \cite{JMLR:v15:vandermaaten14a}, LargeVis \cite{Tang2016a} and most recently presented \textit{Uniform Manifold Approximation and Projection} (UMAP) \cite{McInnes2018a}. In this work, an entropy based outlier score is presented, which can be embedded into the afore mentioned directed graph based dimensionality reduction techniques. The novel outlier score is called \textit{Local Entropy Factor} (LEF) . Compared to other outlier scores it additionally provides detailed information about inlierness. The advantage of the introduced score is, that it is based on the definitions of closeness or distance as used in the dimensionality reduction techniques. Another advantage of the score is that it can be computed directly from the graph constructed during dimensionality reduction. Therefore, no additional graph or model has to be constructed, if a dimensionality reduction is performed on the data. However, the method presented in this work can be used without dimensionality reduction, where it is just constructing the graph. This work also shows how an outlier score presented in another work based on t-SNE \cite{Janssens2012a,Schubert2017a} is embedded into UMAP, called USOS.


Unsupervised outlier detection is a widespread and important research field. Besides the recognition of outliers, identifying inliers might be of interest for some applications. In the validation of autonomous vehicles, both might be of interest in terms of rarely occurring scenarios and in terms of very common scenarios. 

The contributions of this work are as follows. (i) Introduction of the novel unsupervised outlier score LEF based on weighted normalized entropy, which can be applied within graph based dimensionality reduction techniques. (ii) Embedding of a previously defined outlier score into UMAP, leading the score USOS. (iii) Comparing the proposed scores with various state-of-the-art neighborhood based outlier scores on different benchmark datasets. (iv) Applying the scores to a road infrastructure images dataset to validate its capabilities in identifying new traffic scenarios. (v) Providing a publicly available implementation of the LEF in combination with UMAP and the developed openDRIVE \cite{Dupuis2015a} parsing and plotting tool for MATLAB\footnote{Implementations can be found in \url{https://github.com/JWTHI}.}.

The proposed method is more robust with respect to the choice of the number of neighbors, compared to the state-of-the-art scores. This is an important characteristic, since in real world unsupervised problems, the appropriate number of neighbors is not known.

This paper is structured as follows. In Sec. \ref{sec:relWork}, related works in the field of neighborhood based outlier detection are presented. In Sec. \ref{sec:method}, the method itself, including the mathematical background, is introduced. The experiments using the standard datasets, are described in Sec. \ref{sec:exps}. The application to the road infrastructure dataset is shown in \ref{sec:appl}. Sec. \ref{sec:conc} concludes the work.

\section{RELATED WORKS}\label{sec:relWork}
This section focuses on outlier scores which are based on the nearest neighbors of a data point.

In the comprehensive survey \cite{Campos2016a}, various nearest neighborhood based methods are applied to several datasets, providing an empirical analysis. The used datasets are made publicly available to act as a benchmark for outlier detection.

The most basic outlier detection is $k$NN \cite{Ramaswamy2000a}, where the distance to the $k$th nearest neighbor is used as outlier score. In contrast, the extended variant $k$NNW \cite{Angiulli2002a} uses the sum of distances to all $k$ neighbors. The \textit{Outlier Detection using Indegree Number } (ODIN) \cite{Hautamaki2004a} uses the cardinality of the reverse $k$ nearest neighbor sets.

One widely used outlier score is the \textit{Local Outlier Factor} (LOF) \cite{Breunig2000a}. Examining the local reachability density of a point in comparison to the local reachability densities of the point's neighbors, yields the LOF. Several extensions of the LOF have been presented. The \textit{Connectivity-based Outlier Factor} (COF) \cite{Tang2002a} uses connectivity chains to estimate the local reachability density. By using the \textit{Local Outlier Probabilities} (LoOP) \cite{Kriegel2009a}, the density estimates are replaced by the reciprocal of the mean quadratic distance. A more complex replacement of LOF's density estimates can be found in \textit{Local Density Factor} (LDF) \cite{Latecki2007a}, where a variable width Gaussian kernel density estimate is used. The distance used in the LDF are the reachability distances as defined in LOF. An alternative using Gaussian kernels is the \textit{ Kernel Density Estimation Outlier Scores} (KDEOS) \cite{Schubert2014a}, but here the normal distances are used. Another variant is the \textit{Simplified LOF} (SLOF) \cite{Schubert2012a}, where instead of LOF's reachability distance, the distance to the $k$th neighbor is used.

In \cite{Zhang2009a}, the \textit{Local Distance-based Outlier Factor} (LDOF) is presented. The average distance of a point to the $k$ nearest neighbors is divided by the average distance between the neighbors to achieve the LDOF.

In \cite{Kriegel2008a}, the variance of the angles to a point's neighbors is used to determine the outlier score.

The \textit{Local Intrinsic Dimensionality} (LID) uses an estimate of the intrinsic dimensionality to identify outliers in \cite{Houle2018a}.

The following methods are based on definitions used in dimensionality reduction techniques. Using the weighted directed graph constructed in the first stage of t-SNE \cite{Maaten2008a}, the outlier score \textit{Stochastic Outlier Selection} (SOS) was introduced in \cite{Janssens2012a}. Using t-SNE's faster approximation Barnes-Hut-SNE, the outlier scores $k$ \textit{Nearest Neighbor SOS} (KNNSOS) and \textit{Intrinsic SOS} (ISOS) are presented in \cite{Schubert2017a}. The former can be considered as an approximation of the SOS, where in ISOS, the distance measures are transformed based on an intrinsic dimensionality estimate.

The new scores introduced in this work are comparable to the scores which are summarized in this section, in terms of using the neighborhood of a data point to determine the score. However, the principle of calculating the novel score differs from the state-of-the-art scores, since it utilizes the entropy. Though the methods SOS, KNNSOS and ISOS are defined based on a certain dimensionality reduction technique, the introduced method can be applied to any nearest neighbor graph based technique. Additionally, one of the new introduced scores is the UMAP adapted version of KNNSOS.

In \cite{Langner2018a}, the reconstruction error of autoencoders are used to estimate the novelty of traffic scenarios, where in contrast to this work signals over time are used as input. The aim of the application in this work is to detect newly observed infrastructures. A common application of outlier detection in the field of automotive is traffic flow analysis for example in \cite{Djenouri2018c} or \cite{Djenouri2019a}.

\section{METHOD}\label{sec:method}
The novel unsupervised outlier score is based on the directed graph definitions of the dimensionality reduction techniques. The score can be used with any directed nearest neighbor graph based dimensionality reduction technique, however, in this paper, the score is applied in the UMAP and the Barnes-Hut-SNE method. In addition, USOS is introduced, which is the outlier score KNNSOS \cite{Schubert2017a} implemented into UMAP as well. In this section, first the required background of UMAP and Barnes-Hut-SNE is explained.
Next, the entropy based outlier score is introduced for both methods, UMAP and Barnes-Hut-SNE. Then, the application of KNNSOS within UMAP is shown. 

The unsupervised outlier detection is based on an unlabeled dataset $\mathcal{D}=\left\lbrace\bm{x}_1,\dots,\bm{x}_M\right\rbrace$ containing $M$ data points $\bm{x}_m \in \mathbb{R}^N$. The dissimilarity matrix $\bm{D}\in\mathbb{R}_{\geq 0}^{M \times M}$ with an element described by $D_{ij} = d\left(\bm{x}_i,\bm{x}_j\right) $, where $d\left(\ldots\right) \in \mathbb{R}_{\geq0}$ is the dissimilarity measure between the respective data points. In this work, a dataset is represented as a weighted directed graph $G$, where each vertex $v_m$ represents a data point $\bm{x}_m$. An edge $e_{ij}$ from data point $\bm{x}_i$ to $\bm{x}_j$ is weighted  with $P_{ij}$, hence $\bm{P} \in \mathbb{R}^{M \times M}$ contains all weights of the  graph.

\subsection{BACKGROUND}
\subsubsection{UMAP and Barnes-Hut-SNE Directed Graphs}
The focus of this section are the definitions of Barnes-Hut-SNE and UMAP. In both cases the graph construction is realized through point-wise similarity measures. In other words, for each data point a specialized similarity measure for its neighborhood is defined. This way, the constructed graph is going to be directed, since the similarity measures of two points are not necessarily the same. In the following, the construction of the directed graph for both methods is explained. The actual embedding optimizations are not covered in this work, since the outlier scores are just based on the directed graphs in the high dimensional space $\mathbb{R}^N$. Interested readers may refer to the corresponding works for information about the embedding \cite{JMLR:v15:vandermaaten14a}, \cite{McInnes2018a}.

\paragraph{Barnes-Hut-SNE}\label{sec:BHTSNE}
Barnes-Hut-SNE represents the accelerated version of t-SNE, where instead of all data points only the neighbors are taken into consideration for the graph construction. In t-SNE, the similarities are usually denoted as conditional probabilities, nevertheless in this work the terminology ``similarity" is used. Therefore, the similarity between the data points $\bm{x}_i$ and $\bm{x}_j$ using the similarity measure connected to  $\bm{x}_i$ can be stated as
\begin{equation}\label{eq:BHTSNE_sim}
P_{\mathrm{B},ij} = \left\lbrace\begin{array}{cc}
\frac{\exp\left(-d\left(\bm{x}_i,\bm{x}_j\right)^2/2\sigma^2_i\right)}{\sum_{l\in\mathcal{K}_i}\exp\left(-d\left(\bm{x}_i,\bm{x}_l\right)^2/2\sigma^2_i\right)} &  \mathrm{ if } j\in\mathcal{K}_i\\ 
0 & \mathrm{else}
\end{array} 
\right..
\end{equation}
With $\mathcal{K}_i$ containing the indices of the $k$ nearest neighbors of $i$ and $\sigma^2_i$ the variance of a Gaussian. The value of  $\sigma^2_i$ is determined by a binary search, such that the perplexity $u$ is
\begin{equation}
u = 2^{-\sum_{j \in \mathcal{K}_i} P_{\mathrm{B},ij}\log_2 P_{\mathrm{B},ij}},
\end{equation}
where the number of neighbors is set to $k=\left\lfloor 3u\right\rfloor$ here, as in \cite{JMLR:v15:vandermaaten14a}. The $\lfloor \ldots\rfloor$ is the floor operator. This way, the sparse weight Matrix $\bm{P}_\mathrm{B}\in\mathbb{R}^{M \times M}$ is constructed.

\paragraph{UMAP}
The UMAP method is derived from topology theory and is a manifold learning technique. As in Barnes-Hut-SNE, a directed weighted graph is constructed using only the $k$ nearest neighbors of the data points. The similarity between $\bm{x}_i$ and $\bm{x}_j$ is calculated as
\begin{equation}\label{eq:UMAP_sim}
P_{\mathrm{U},ij} = \left\lbrace\begin{array}{cc}
\exp\left(\frac{-\max(0,d\left(\bm{x}_i,\bm{x}_j\right)-\rho_i)}{\sigma_i}\right) &  \mathrm{ if } j\in\mathcal{K}_i\\ 
0 & \mathrm{else}
\end{array} 
\right..
\end{equation}
The dissimilarity to the closest neighbor of $\bm{x}_i$ is denoted by $\rho_i$. By subtracting the dissimilarity of the most similar neighbor, it will have similarity of one. However, this definition is required to ensure a local connectivity of the approximated manifold. As in the Barnes-Hut-SNE method, the $\sigma_i$ is determined such that it fulfills a certain criterion, here
\begin{equation}\label{eq:UMAP_BinSearch}
\sum_{j\in\mathcal{K}_i} \exp\left(\frac{-\max(0,d\left(\bm{x}_i,\bm{x}_j\right)-\rho_i)}{\sigma_i}\right) = \log_2(k).
\end{equation}
Given the definitions in Eq. (\ref{eq:UMAP_sim}) and (\ref{eq:UMAP_BinSearch}), the sparse weight matrix $\bm{P}_\mathrm{U}\in\mathbb{R}^{M \times M}$ of UMAP's $k$ nearest neighbor graph is calculated.

\subsubsection{KNNSOS}
In this section, the outlier score KNNSOS \cite{Schubert2017a} is briefly summarized. The initial score SOS \cite{Janssens2012a} is defined using the term binding probabilities instead of t-SNE's conditional probabilities. An outlier is defined as a data point which is frequently not probable to be linked. In SOS, all the data points are used to determine the score, where in KNNSOS only the reverse $k$ nearest neighbor set of a point are used. Hence, only the incoming edges are considered. Therefore, the outlier score of $\bm{x}_i$ is defined as
\begin{align}
\text{KNNSOS}_i &= \prod_{j\neq i} 1-P_{\mathrm{B},ji},
\end{align}
with $P_{\mathrm{B},ji}$ calculated as in Eq. (\ref{eq:BHTSNE_sim}). The score leads to high values for data points being weakly linked to other data points. If no link exists at all, the score will be $1$. In other words, the data point $\bm{x}_i$ is not a member of any neighborhood.

\subsection{Entropy Based Outlier Detection}
The novel entropy based outlier detection is presented in this section. It can be used in any of the graph based dimensionality reduction techniques. This score is designed to contain information about inlierness in addition to identifying local outliers. Hence, first the definition of inlierness is required. Here the membership strength of an inlier is defined as maximal if it is the nearest neighbor to all its $k$ neighbors and minimal if it is not in the neighborhood of any of its neighbors. Additionally, the score should be higher when the incoming similarities are equally distributed. Moreover, the value has to be weighted by the sum of incoming similarities, such that a point being nearest neighbor to all its neighbors gets a higher score than a point being an equally weak neighbor to all its neighbors. This definition favors data points which have the same similarities to all their neighbors over varying similarities. The entropy based score is meant to detect outlying points within their neighborhood. Hence, big outlying groups will not be detected as outliers, since they form a neighborhood.

The Shannon entropy fulfills the equally distributed requirement of the above definition. Therefore, a weighted normalized entropy is utilized in this work as the measure of inlierness. Identifying local outliers is achieved by selecting the data points with a low inlierness value. Let the relative incoming similarities $\hat{\bm{P}}$ be defined as
\begin{equation}
\hat{P}_{ji} = \frac{P'_{\mathrm{U},ji}}{\sum_{j\in\mathcal{K}_i}P'_{\mathrm{U},ji}},
\end{equation}
with $P'_{\mathrm{U},ij} = P_{\mathrm{U},ij}/\log_2(k)$ such that the outgoing similarities sum to one, see Eq. (\ref{eq:UMAP_BinSearch}).
The weighted normalized entropy is calculated as
\begin{equation}
\text{USLEF}_i = -\frac{\sum_{j\in\mathcal{K}_i}P'_{\mathrm{U},ji}}{k}\sum_{j\in\mathcal{K}_i}\hat{P}_{ji}\log_2\left(\hat{P}_{ji}\right).
\end{equation}
By using the relative incoming similarities, it is ensured that they sum up to one and hence the normalized entropy is bound to one. The normalized entropy of the relative incoming similarities would lead to one for data points with equal incoming similarities, independent of the actual value of the similarities. Therefore, as described before, the normalized entropy is weighted with the sum of incoming similarities divided by the maximum possible value $k$. The acronym USLEF stands for UMAP-Sparse LEF.

Besides the USLEF, its non sparse version, which will be called ULEF, is also used in this work. For this purpose, the local similarity measures of UMAP are extended for non neighborhood points. This extension collides with the definition of UMAP that the local similarity measures are just defined within the neighborhood. ULEF is calculated as USLEF but using the incoming similarities of not only the neighbors with incoming edges but all its $k$ neighbors.

As mentioned above, the score can be used with the similarities as defined in Barnes-Hut-SNE as well. The sparse version tSLEF utilizes the $\bm{P}_{\mathrm{B}}$ instead of the $\bm{P}'_{\mathrm{U}}$. The non sparse version is called tLEF, where as before, the incoming similarities of all neighbors are calculated independent of the actual point being in the neighborhood.

\subsection{KNNSOS with UMAP}
The KNNSOS score \cite{Schubert2017a} was defined within Barnes-Hut-SNE. In this work, its potential using UMAP is shown. The application as such is straight forward. But, the similarities generated by  UMAP need to be adjusted. This is due to the local connectivity requirement of UMAP, which results in the nearest neighbor having a similarity of one. Each data point which is the nearest neighbor of any other point would have an outlier score of zero, no matter what the other similarities are. Therefore, the weights of the graph are adjusted by using $P'_{\mathrm{U},ij}$ as defined above.

Provided with the adjusted weights, the application of KNNSOS in UMAP is given by the outlier score USOS, defined as
\begin{equation}
\text{USOS}_i = \prod_{j\neq i} 1-P'_{\mathrm{U},ji}.
\end{equation}
Since the weights are adjusted, the interpretation of the outlier score changes. The smallest possible value which can be achieved is $\text{USOS}_i = (1-1\log_2(k))^{\vert\mathcal{RK}_i\vert}$, with $\mathcal{RK}_i$ being the set of reverse $k$ nearest neighbors. Hence, the smallest possible number varies for each data point. This may need to be considered in data analysis steps. In this work, the score wont be adjusted, since if more incoming edges exist, the data point can be considered less an outlier than others, which consist of less incoming edges. This characteristic is used in ODIN as well. If UMAP is used as dimensionality reduction technique, the USOS can be determined with little effort, since the graph is already constructed. Besides this advantage, using the same definitions of similarity as used in the projection technique might aid towards interpretability.

\section{EXPERIMENTS}\label{sec:exps}
The presented scores ULEF, USLEF, USOS, tLEF and tSLEF are investigated with respect to their outlier detection capabilities in this section. Therefore, they are applied to several real world datasets and compared to state-of-the-art outlier scores. This section provides a brief summary of the used datasets and the evaluation measure. The results are shown and discussed in this section as well. The implementation of the UMAP based scores uses parts of the publicly available implementation \cite{McInnes2018b}, such as performing the binary search. The Barnes-Hut-SNE based scores are a modified version of \cite{Janssens2012b}.

\subsection{Datasets}
The datasets for the experiments in this section were taken from \cite{Campos2016b}. All datasets were normalized and do not contain duplicates. Furthermore, in case of categorical attributes, the \textit{Inverse Document Frequency} (IDF) coded versions are used. No further processing on the downloaded datasets is applied. In this work only the standard versions of the datasets are used. Let the number of samples per dataset be $M$, the number of outliers $M_\mathrm{O}$.

\subsection{Evaluation Method}
The evaluation is performed in such a way, that each outlier point is used individually with all the inliers. Hence, the dataset Glass yields 9 individual outlier datasets $\mathcal{D}_{\mathcal{O},1}\dots\mathcal{D}_{\mathcal{O},9}$ each containing 205 instances and one outlier. This differs from the evaluation  performed in \cite{Campos2016a}, as in this work the focus is set to be based on the local structure.

Each of the outlier datasets is then processed with different outlier scores, including the ones introduced in this work. For this purpose, the number of neighbors $k$ is varied from 3 to 100. The $k$ is selected larger or equal to 3 such that the perplexity in Barnes-Hut-SNE is at least 1 (see Sec. \ref{sec:BHTSNE}). For each $k$, an evaluation measure is calculated. Here the ROC AUC\footnote{\textit{ Area Under Curve} (AUC) of the \textit{Receiver Operating Characteristic} (ROC) curve.} as used in \cite{Campos2016a}
\begin{equation}
AUC_k =\frac{1}{\vert\mathcal{I}\vert}\sum_{i\in\mathcal{I}}{\left\lbrace\begin{array}{cc}
	1 & \text{if}f(o)> f(i) \\ 
	0.5 & \text{if}f(o)= f(i) \\ 
	0 & \text{if}f(o)< f(i)
	\end{array} 
	\right.}
\end{equation}
is utilized, where here $o$ is the outlier and $\mathcal{I}$ the set of inliers of the given outlier dataset, with its cardinality $\vert\mathcal{I}\vert$ . The scores are represented by $f(\ldots)$. An $AUC_k$ of 1 indicates that the outlier value has the highest outlier score of all data points. Whereas, a value of 0 indicates that the outlier has the lowest value and hence is not detected as such.

The maximum and the average values over all $k$s of $AUC_k$ are determined per outlier dataset. Then, the averages of both scores per dataset are determined as
\begin{align}
\overline{AUC}_\mathrm{max} (\mathcal{D}) &= \frac{1}{M_\mathrm{O}}\sum_{i=1}^{M_\mathrm{O}}\max_{k=3,\dots,100}\left({AUC_k(\mathcal{D}_{\mathcal{O},i})}\right)\\
\overline{AUC}_\mathrm{avg} (\mathcal{D}) &= \frac{1}{M_\mathrm{O}}\sum_{i=1}^{M_\mathrm{O}}\frac{1}{98}\sum_{k=3}^{100}{AUC_k(\mathcal{D}_{\mathcal{O},i})}.
\end{align}
Those values can be considered as indicators for maximum possible and average performance given a certain dataset. In a final step, both values are averaged for all datasets.

\subsection{Results}
\begin{table*}
	\vspace{2mm}
	\caption{$\overline{AUC}_\mathrm{max}$ per dataset and method}
	\label{tb:maxTable}
	\centering
	\resizebox{.95\textwidth}!{\begin{tabular}{|c|r|r|r|r|r|r|r|r|r|r|r|r|r|}
			\hline  
			Dataset & ULEF & USLEF & USOS & tLEF & tSLEF & KNNSOS & ISOS & KNN & KNNW & LOF & LDOF & LoOP & COF\\ \hline
			Annthyroid 	& $.78 \pm .24 $  & $.78 \pm .24 $  & $\mathbf{.80 \pm .23 }$  & $.79 \pm .24 $  & $.79 \pm .24 $  & $\mathbf{.80 \pm .23 }$  & $.79 \pm .24 $  & $.67 \pm .26 $  & $.68 \pm .26 $  & $.74 \pm .26 $  & $.78 \pm .23 $  & $.77 \pm .24 $  & $.73 \pm .27 $ \\ 
			Arrhythmia & $.76 \pm .25 $  & $\mathbf{.76 \pm .24 }$  & $.76 \pm .25 $  &  $.76 \pm .25 $ & $\mathbf{.76 \pm .24} $   & $.76 \pm .25 $  & $.76 \pm .25 $  & $.76 \pm .27 $  & $.76 \pm .26 $  & $.75 \pm .27 $  & $.76 \pm .25 $  & $.76 \pm .25 $  & $.75 \pm .25 $ \\ 
			Cardiotocography & $.86 \pm .19 $  & $.85 \pm .19 $  & $\mathbf{.87 \pm .18 }$   & $.85 \pm .19 $ & $.86 \pm .19 $ & $.87 \pm .19 $  & $.86 \pm .19 $  & $.80 \pm .23 $  & $.82 \pm .22 $  & $.83 \pm .21 $  & $.83 \pm .22 $  & $.86 \pm .21 $  & $.84 \pm .23 $ \\ 
			Glass & $.85 \pm .11 $  & $.83 \pm .12 $  & $.83 \pm .10 $  &  $.80 \pm .13 $ & $.80 \pm .12 $ & $.79 \pm .16 $  & $.81 \pm .12 $  & $.88 \pm .06 $  & $.89 \pm .06 $  & $\mathbf{.92 \pm .06 }$  & $.84 \pm .07 $  & $.87 \pm .08 $  & $.89 \pm .11 $ \\ 
			HeartDisease & $.78 \pm .20 $  & $.78 \pm .20 $  & $.77 \pm .21 $  & $.79 \pm .20 $  & $.79 \pm .20 $  & $.78 \pm .20 $  & $.77 \pm .21 $  & $\mathbf{.85 \pm .18 }$  & $.84 \pm .18 $  & $.84 \pm .18 $  & $.82 \pm .20 $  & $.82 \pm .20 $  & $\mathbf{.85 \pm .18 }$ \\ 
			Hepatitis & $.72 \pm .19 $  & $.73 \pm .21 $  & $.68 \pm .19 $  &  $.71 \pm .20 $ & $.73 \pm .19 $   & $.65 \pm .21 $  & $.66 \pm .20 $  & $.82 \pm .11 $  & $.79 \pm .15 $  & $\mathbf{.82 \pm .19 }$  & $.79 \pm .14 $  & $.79 \pm .15 $  & $.82 \pm .14 $ \\ 
			InternetAds & $.92 \pm .20 $  & $.92 \pm .16 $  & $.92 \pm .18 $  &  $.92 \pm .18 $  & $.92 \pm .16 $  & $.91 \pm .20 $  & $\mathbf{.94 \pm .15 }$  & $.86 \pm .24 $  & $.87 \pm .23 $  & $.92 \pm .18 $  & $.92 \pm .19 $  & $.92 \pm .19 $  & $.91 \pm .19 $ \\ 
			Ionosphere & $.95 \pm .10 $  & $.95 \pm .10 $  & $.95 \pm .11 $  & $.95 \pm .10 $  & $.95 \pm .10 $  & $.95 \pm .11 $  & $.95 \pm .11 $  & $\mathbf{.97 \pm .07 }$  & $.97 \pm .08 $  & $.95 \pm .09 $  & $.93 \pm .12 $  & $.95 \pm .09 $  & $\mathbf{.97 \pm .07 }$ \\ 
			Lymphography & $\mathbf{1.00 \pm .00 }$  & $1.00 \pm .01 $  & $1.00 \pm .01 $  & $1.00 \pm .01 $  & $1.00 \pm .01 $  & $1.00 \pm .01 $  & $1.00 \pm .01 $  & $\mathbf{1.00 \pm .00 }$  & $\mathbf{1.00 \pm .00 }$  & $\mathbf{1.00 \pm .00 }$  & $\mathbf{1.00 \pm .00 }$  & $\mathbf{1.00 \pm .00} $  & $\mathbf{1.00 \pm .00 }$ \\ 
			PageBlocks & $.93 \pm .13 $  & $.93 \pm .13 $  & $.92 \pm .15 $   & $.92 \pm .13 $ & $.93 \pm .13 $  & $.92 \pm .14 $  & $.93 \pm .13 $  & $.92 \pm .14 $  & $.93 \pm .14 $  & $.95 \pm .10 $  & $\mathbf{.96 \pm .06 }$  & $.94 \pm .09 $  & $.91 \pm .18 $ \\ 
			Parkinson & $.95 \pm .10 $  & $.94 \pm .11 $  & $.95 \pm .09 $   & $.95 \pm .09 $ & $.94 \pm .09 $ & $\mathbf{.97 \pm .08 }$  & $.95 \pm .08 $  & $.89 \pm .13 $  & $.92 \pm .11 $  & $.93 \pm .16 $  & $.85 \pm .21 $  & $.94 \pm .12 $  & $.93 \pm .11 $ \\ 
			PenDigits & $\mathbf{1.00 \pm .00 }$  & $1.00 \pm .01 $  & $1.00 \pm .01 $  & $1.00 \pm .01 $  & $1.00 \pm .01 $  & $.99 \pm .01 $  & $.99 \pm .01 $  & $\mathbf{1.00 \pm .00 }$  & $\mathbf{1.00 \pm .00 }$  & $1.00 \pm .01 $  & $.95 \pm .08 $  & $\mathbf{1.00 \pm .00 }$  & $1.00 \pm .01 $ \\ 
			Pima & $.68 \pm .27 $  & $.68 \pm .27 $  & $.68 \pm .26 $  & $.67 \pm .27 $  & $.67 \pm .27 $  & $.67 \pm .27 $  & $.67 \pm .27 $  & $.76 \pm .22 $  & $.75 \pm .22 $  & $.74 \pm .20 $  & $.67 \pm .27 $  & $.69 \pm .26 $  & $\mathbf{.78 \pm .21 }$ \\ 
			Shuttle & $.98 \pm .01 $  & $.98 \pm .02 $  & $.98 \pm .01 $  & $.98 \pm .01 $  & $.98 \pm .01 $  & $.98 \pm .02 $  & $\mathbf{.99 \pm .01 }$  & $.96 \pm .03 $  & $.97 \pm .02 $  & $.98 \pm .02 $  & $.98 \pm .02 $  & $\mathbf{.99 \pm .01 }$  & $.98 \pm .02 $ \\ 
			SpamBase & $\mathbf{.82 \pm .21 }$  & $.81 \pm .23 $  & $.80 \pm .24 $  & $.82 \pm .22 $  & $.82 \pm .22 $  & $.80 \pm .24 $  & $.80 \pm .24 $  & $.73 \pm .22 $  & $.75 \pm .22 $  & $.77 \pm .21 $  & $.78 \pm .20 $  & $.78 \pm .21 $  & $.75 \pm .20 $ \\ 
			Stamps & $.95 \pm .04 $  & $.95 \pm .05 $  & $.90 \pm .07 $   & $.94 \pm .06 $ & $.95 \pm .06 $  & $.90 \pm .08 $  & $.92 \pm .06 $  & $.95 \pm .03 $  & $\mathbf{.96 \pm .02 }$  & $.94 \pm .04 $  & $.93 \pm .03 $  & $.94 \pm .04 $  & $.95 \pm .04 $ \\ 
			WBC & $.93 \pm .08 $  & $.93 \pm .08 $  & $.94 \pm .09 $  & $.93 \pm .09 $ & $.93 \pm .08 $  & $.93 \pm .11 $  & $.94 \pm .09 $  & $\mathbf{.99 \pm .02 }$  & $\mathbf{.99 \pm .02}$  & $\mathbf{.99 \pm .02 }$  & $.98 \pm .03 $  & $.98 \pm .03 $  & $.98 \pm .02 $ \\ 
			WDBC & $.97 \pm .06 $  & $\mathbf{.97 \pm .05}$  & $\mathbf{.97 \pm .05}$   & $\mathbf{.97 \pm .05}$ & $.97 \pm .06 $  & $\mathbf{.97 \pm .05 }$  & $\mathbf{.97 \pm .05} $  & $.95 \pm .08 $  & $.95 \pm .08 $  & $\mathbf{.97 \pm .05}$  & $\mathbf{.97 \pm .05 }$  & $\mathbf{.97 \pm .05}$  & $.95 \pm .07 $ \\ 
			WPBC & $.53 \pm .24 $  & $.53 \pm .27 $  & $.52 \pm .25 $  & $.52 \pm .24 $  & $.52 \pm .24 $  & $.53 \pm .28 $  & $.51 \pm .24 $  & $.57 \pm .23 $  & $.56 \pm .23 $  & $.55 \pm .23 $  & $.55 \pm .23 $  & $.53 \pm .24 $  & $\mathbf{.59 \pm .25 }$ \\ 
			Waveform & $.76 \pm .27 $  & $.76 \pm .27 $  & $.76 \pm .27 $  & $.76 \pm .28 $ & $.76 \pm .27 $  & $.75 \pm .27 $  & $.76 \pm .27 $  & $\mathbf{.78 \pm .24 }$  & $.78 \pm .25 $  & $.77 \pm .26 $  & $.78 \pm .24 $  & $.77 \pm .27 $  & $.73 \pm .27 $ \\ 
			Wilt & $.80 \pm .19 $  & $.81 \pm .20 $  & $.83 \pm .20 $   & $.83 \pm .18 $ & $.82 \pm .19 $  & $\mathbf{.84 \pm .17 }$  & $.83 \pm .18 $  & $.58 \pm .19 $  & $.61 \pm .19 $  & $.72 \pm .26 $  & $.83 \pm .15 $  & $.80 \pm .20 $  & $.74 \pm .23 $ \\ 
			\hline 
	\end{tabular}}
\vspace{-2mm}
\end{table*}

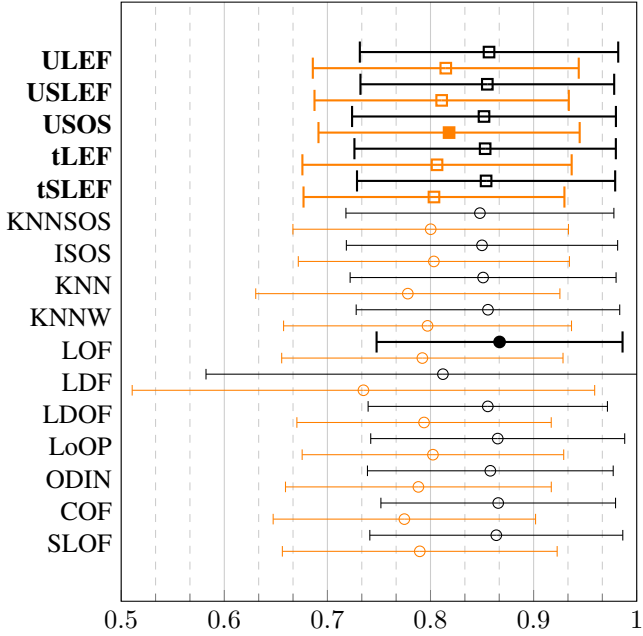
\begin{figure}
	\centering
			\begin{tikzpicture}
		\begin{axis}%
			[
			xmin =0.5,
			xmax =1.0,
			xtick pos=left,
			ytick style={draw=none},
			ytick={1.5,3.5,5.5,7.5,9.5,11.5,13.5,15.5,17.5,19.5,21.5,23.5,25.5,27.5,29.5,31.5,33.5},
			yticklabels={SLOF,COF,ODIN,LoOP,LDOF,LDF,LOF,KNNW,KNN,ISOS,KNNSOS,\textbf{tSLEF},\textbf{tLEF},\textbf{USOS},\textbf{USLEF},\textbf{ULEF}},
			y post scale=1.4,
			minor tick num=2,
			xmajorgrids,
			xminorgrids
			]
			\addplot [color=black, only marks,mark=o,]
			plot [error bars/.cd, x dir = both, x explicit,]
			table[x =x, y =y, x error = ex]{maxMean.dat};\label{pl:maxMean1}
			\addplot [color=orange, only marks,mark=o,]
			plot [error bars/.cd, x dir = both, x explicit,]
			table[x =x, y =y, x error = ex]{meanMean.dat};\label{pl:meanMean1}
			
			\addplot [color=black, only marks,mark=*, thick]
			plot [error bars/.cd, x dir = both, x explicit, error bar style={line width=0.8pt,solid},
			error mark options={line width=0.8pt,mark size=4pt,rotate=90},]
			table[x =x, y =y, x error = ex]{maxMeanBest.dat};\label{pl:maxMean2}
			\addplot [color=orange, only marks,mark=square*, thick]
			plot [error bars/.cd, x dir = both, x explicit, error bar style={line width=0.8pt,solid},
			error mark options={line width=0.8pt,mark size=4pt,rotate=90},]
			table[x =x, y =y, x error = ex]{meanMeanBest.dat};\label{pl:meanMean2}
			
			\addplot [color=black, only marks,mark=square, thick]
			plot [error bars/.cd, x dir = both, x explicit, error bar style={line width=0.8pt,solid},
			error mark options={line width=0.8pt,mark size=4pt,rotate=90},]
			table[x =x, y =y, x error = ex]{maxMeanOwn.dat};\label{pl:maxMean3}
			\addplot [color=orange, only marks,mark=square, thick]
			plot [error bars/.cd, x dir = both, x explicit, error bar style={line width=0.8pt,solid},
			error mark options={line width=0.8pt,mark size=4pt,rotate=90},]
			table[x =x, y =y, x error = ex]{meanMeanOwn.dat};\label{pl:meanMean3}
		\end{axis}
	\end{tikzpicture}
	\caption{Performance per method: (\ref{pl:maxMean3}/\ref{pl:maxMean1}/\ref{pl:maxMean2}) average $\overline{AUC}_\mathrm{max}$ and (\ref{pl:meanMean3}/\ref{pl:meanMean1}/\ref{pl:meanMean2}) average $\overline{AUC}_\mathrm{max}$ over all datasets. The squares indicate scores introduced in this work and the circles other scores.}
	\label{fig:maxmeanMeanErrorBar}
\end{figure}
The application and evaluation of the outlier scores to the various datasets as described in the former sections are discussed here. The average of all datasets for the max and the mean characteristics are shown in Fig. \ref{fig:maxmeanMeanErrorBar}. The results of the scores introduced in this work are drawn thick and the square is an indicator of the average value. The remaining methods are depicted normal and with a circle. The whiskers in all cases indicate the standard deviation. The best performing score is indicated thick and with a filled marker.

From the plots shown in Fig. \ref{fig:maxmeanMeanErrorBar}, it can be concluded that over all, the introduced scores are at a comparable level in terms of maximum performance (\ref{pl:maxMean3}/\ref{pl:maxMean1}/\ref{pl:maxMean2}). However, LOF performed better than the rest. In terms of the average values, the introduced scores, especially the ones in combination with UMAP (ULEF, USLEF, USOS) show the best performance over all scores. They offer a higher robustness against variations of $k$, as it becomes clear from Fig. \ref{fig:maxmeanMeanErrorBar}, where the average values over all $k$s are shown (\ref{pl:meanMean3}/\ref{pl:meanMean1}/\ref{pl:meanMean2}). Hence, for a randomly chosen $k$, the methods with the lowest risk are ULEF and USOS. This is an essential attribute, since selecting an appropriate $k$ is difficult because the ground truth is not available for real world unsupervised tasks.

The scores ULEF and tLEF and their sparse versions USLEF and tSLEF are at a comparable level for the maximum values. But, the non-sparse versions are slightly better for the average values $\overline{AUC}_\mathrm{avg}$ (\ref{pl:meanMean3}/\ref{pl:meanMean1}/\ref{pl:meanMean2}). The non-sparse versions are causing additional computational cost, since the missing incoming edges are calculated as well. Therefore, deciding between the sparse and the non-sparse versions can be considered as tradeoff between computational cost and accuracy.

A more detailed list of the different maximum values is depicted in Tb. \ref{tb:maxTable}. To ease readability, only the best performing state-of-the-art scores are listed. As with the other scores, the introduced scores may suit for certain datasets better than for others. There is no score that is generally outperforming the others. For example, LOF is outperforming the ULEF on 8 datasets, whereas ULEF is outperforming LOF on 7. Accordingly, in terms of maximum accuracy, the LOF is outperforming the ULEF, as already depicted in Fig. \ref{fig:maxmeanMeanErrorBar} (\ref{pl:maxMean3}/\ref{pl:maxMean1}/\ref{pl:maxMean2}). On the other hand, the ULEF is outperforming the LOF in terms of average accuracy, thereby depicting its robustness against the number of neighbors $k$ (see Fig. \ref{fig:maxmeanMeanErrorBar} (\ref{pl:meanMean3}/\ref{pl:meanMean1}/\ref{pl:meanMean2})).

\section{Application to Road Infrastructure Images}\label{sec:appl}
The application of the introduced scores to road infrastructure images is presented in this section. Furthermore, the construction of the image dataset is explained here. The experiments are evaluated for various scores, such that an application based analysis can be provided.
\begin{figure}
	\vspace{2mm}
	\centering
	\includegraphics[width=0.95\linewidth]{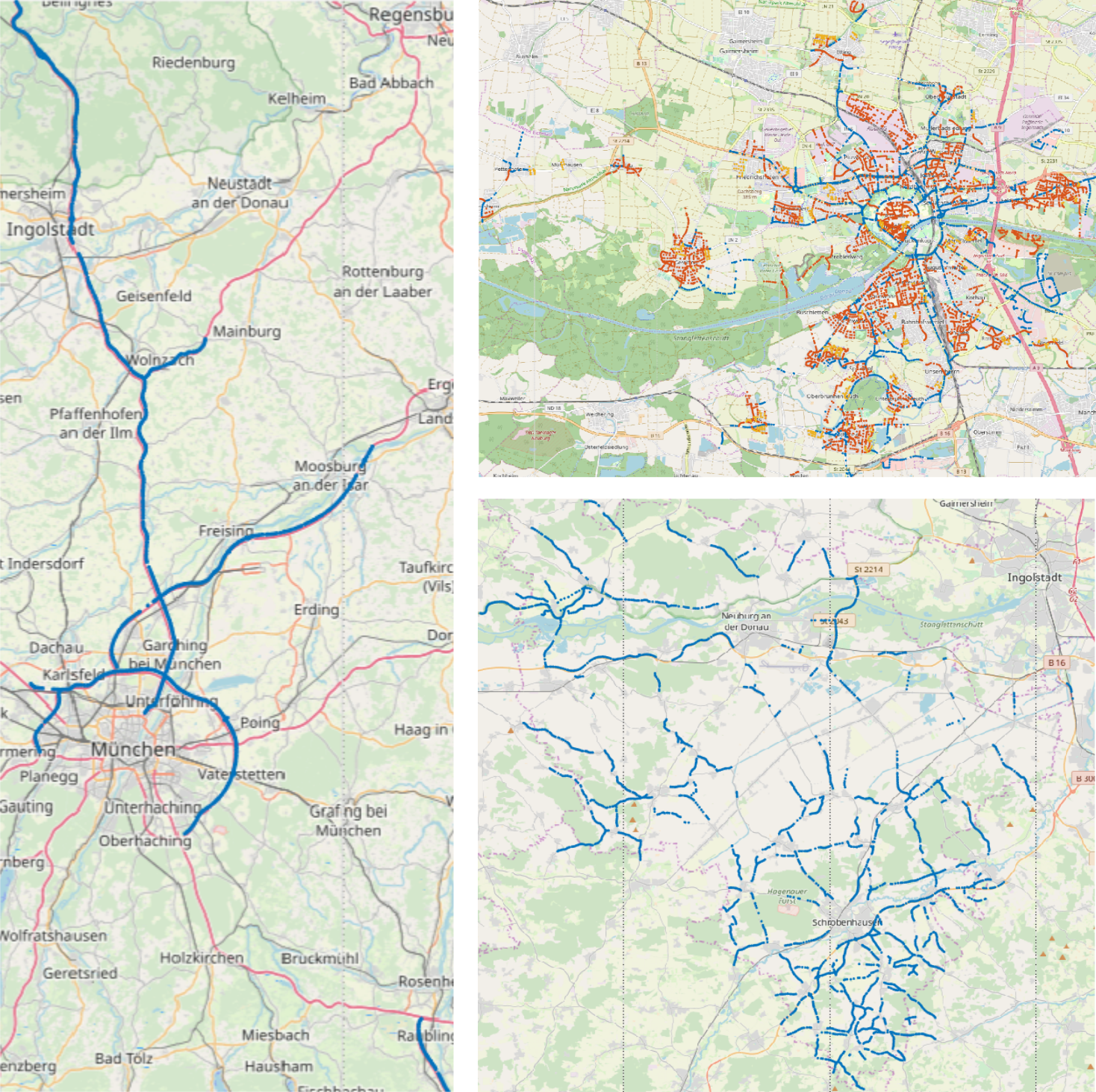}
	\caption{Centers of Images. Classes: Left (i); Right Bottom (ii); Right Top (iii) blue, (iv) red, (v) orange. \cite{OpenStreetMap}}
	\label{fig:NodesOnMap}
	\vspace{-2mm}
\end{figure}

As already mentioned above, identifying newly observed traffic scenarios is a key aspect for the validation process of autonomous vehicles. The road infrastructure builds a crucial part of a scene. The results of this work are thought to serve as a case study. Here a dataset of highway infrastructure images is assumed as pre-recorded dataset. Various other infrastructure classes are compared in terms of outlierness to the highway dataset. It is important to note, that the used classes are just defined to validate the method, since it can be assumed that an highway should be differentiable from the most other infrastructures images. Hence, the task is not to differentiate between highway and not highway, but the task is to differentiate between unknown and known infrastructure images. To the best knowledge of the authors, this is the first work in which nearest neighbor based outlier detection is applied to road infrastructure images. 

\subsection{Data Generation}
The dataset generation used for this application is part of this work. Five different infrastructure classes with a total number of 21\,985 images are generated. The classes are (i) highway, (ii) rural roads and inner-city roads with a speed limit of (iii) \unit[50]{km/h}, (iv) \unit[30]{km/h} and (v) \unit[5]{km/h}. The detailed steps are described below.

The images are generated from map data which is provided by OpenStreetMap \cite{OpenStreetMap}. In a first step, for a given geographic area, all nodes\footnote{Nodes in OpenStreetMap define a single position on the map. And a way consists of an ordered list of Nodes, defining the shape of the road.} which fit the class criteria are extracted using the Overpass API. All the data is extracted from within and around the city of Ingolstadt in Germany.
\begin{table}
	\caption{Road Infrastructure Dataset}\label{tb:roadData}
	\centering
	\resizebox{0.95\columnwidth}{!}{%
		\begin{tabular}{l|c|c|c}
			Class & Search Area & Search Criteria & Nodes\\
			\hline
			(i) Highway & Upper Bavaria &  A9, A92, A93, A99 & 5\,447\\
			\hline
			(ii) Rural & Neub. Schrobenh. & 70, 80, 90, \unit[100]{km/h} & 5\,636\\
			\hline
			(iii) City 50 & Ingolstadt & \unit[50]{km/h} & 4\,604\\
			\hline
			(iv) City 30 & Ingolstadt & \unit[30]{km/h} & 5\,152\\
			\hline
			(v) City 5 & Ingolstadt & Living Street & 1\,146\\
	\end{tabular}}
\end{table}

The highway nodes are extracted by using upper Bavaria as overall search area and the highways A9, A92, A93, and A99. The district of Neuburg Schrobenhausen, which borders Ingolstadt, is used to extract the rural road nodes. There, all nodes which have a speed limit of either \unit[100]{km/h}, \unit[90]{km/h}, \unit[80]{km/h} or \unit[70]{km/h} are considered. For the inner city nodes, Ingolstadt is selected as search area. For the classes (iii) and (iv), the corresponding speed limits are used as filters. In the case of class (v), the road type is filtered for living streets. The various settings as well as the number of resulting nodes is summarized in Tb. \ref{tb:roadData}. The positions of the nodes are visualized in Fig. \ref{fig:NodesOnMap}. 
\begin{figure}
	\vspace{2mm}
	\centering
	\noindent
	\resizebox{0.95\columnwidth}{!}{%
		\noindent
		\setlength{\tabcolsep}{2pt}
		\newlength{\mywidth}
		\setlength{\mywidth}{64pt}
		\begin{tabular}{c c c c c}
			(i) & (ii) & (iii) & (iv) & (v)\\
			\includegraphics[width=\mywidth,height=!]{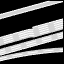}&
			\includegraphics[width=\mywidth,height=!]{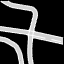}&
			\includegraphics[width=\mywidth,height=!]{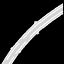}&
			\includegraphics[width=\mywidth,height=!]{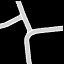}&
			\includegraphics[width=\mywidth,height=!]{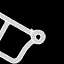}\\
			\includegraphics[width=\mywidth,height=!]{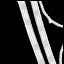}&
			\includegraphics[width=\mywidth,height=!]{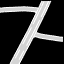}&
			\includegraphics[width=\mywidth,height=!]{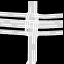}&
			\includegraphics[width=\mywidth,height=!]{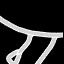}&
			\includegraphics[width=\mywidth,height=!]{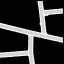}\\
		\end{tabular}%
	}
	\caption{Example images of the different road infrastructure classes (Tb. \ref{tb:roadData}).}\label{fig:roadExamples}
	\vspace{-2mm}
\end{figure}

In a next step, each node is considered as the center of a bounding box with the size of $\unit[100]{m}\times\unit[100]{m}$. The corresponding maps are downloaded and then converted into the OpenDRIVE format \cite{Dupuis2015a} using the NETCONVERT tool of SUMO \cite{SUMO2018}. As part of this work, a tool is developed to generate adjustable images given OpenDRIVE maps as input. In this work, the image generation is adjusted such that all non drivable lanes (biking, sidewalk, restricted, rail) are ignored and the images are of size $64\times64$ pixels. The lane surface itself is colored in gray, the lane markings are given in white and the remaining area is set to black. In Fig. \ref{fig:roadExamples} some example images for the different classes are depicted.

\subsection{Results}
The evaluation of the various outlier scores follows the same intuition as in the general experiments. Hence, the outlier detection remains unsupervised, only the evaluation is based on the classes as defined above. Here, the highway class is considered to be the respective inlier dataset. All the remaining classes are considered to be outliers. In terms of the application, this can be thought of having only data collected on a highway, then gathering data on other infrastructures and comparing their novelty based on the pre-recorded data. As described in Sec. \ref{sec:exps}, each image of the outlier classes is used separately to evaluate the outlier score compared to the highway images.

\begin{figure}[t]
\vspace{2mm}
	\begin{tabular}{c c }
		(ii) Rural&(iii) City 50\\
		\begin{tikzpicture}
\footnotesize
\begin{axis}[
grid=major,
width=0.47\columnwidth,
height=0.5\columnwidth,
xlabel=$k$,
ylabel=$\overline{AUC}$,
ymin=0.3, 
ymax=0.9,
xmin=0,
xmax=100,
xticklabels={,0,20,40,60,80,100}]

\addplot[draw=none,color=black,mark=square*] table [x=k,y=LDOF]{HighvsRural.dat};\label{pl:LDOF}
\addplot[draw=none,color=red,mark=square]  table [x=k,y=USOS]{HighvsRural.dat};\label{pl:USOS}
\addplot[draw=none,color=blue,mark=o] table [x=k,y=tLEF]{HighvsRural.dat};\label{pl:tLEF}
\addplot[draw=none,color=blue,mark=square*] table [x=k,y=SOS]{HighvsRural.dat};\label{pl:KNNSOS}
\addplot[draw=none,color=red,mark=*] table [x=k,y=ULEF]{HighvsRural.dat};\label{pl:ULEF}

\end{axis}
\end{tikzpicture}&
		\begin{tikzpicture}
\footnotesize
\begin{axis}[
grid=major,
width=0.47\columnwidth,
height=0.5\columnwidth,
xlabel=$k$,
ylabel=$\overline{AUC}$,
ymin=0.3, 
ymax=0.9,
xmin=0,
xmax=100,
xticklabels={,0,20,40,60,80,100}]

\addplot[draw=none,color=black,mark=square*] table [x=k,y=LDOF]{Highvs50.dat};\label{pl:LDOF}
\addplot[draw=none,color=red,mark=square]  table [x=k,y=USOS]{Highvs50.dat};\label{pl:USOS}
\addplot[draw=none,color=blue,mark=o] table [x=k,y=tLEF]{Highvs50.dat};\label{pl:tLEF}
\addplot[draw=none,color=blue,mark=square*] table [x=k,y=SOS]{Highvs50.dat};\label{pl:KNNSOS}
\addplot[draw=none,color=red,mark=*] table [x=k,y=ULEF]{Highvs50.dat};\label{pl:ULEF}

\end{axis}
\end{tikzpicture}\\
		(iv) City 30 & (v) City 5\\
		\begin{tikzpicture}
\footnotesize
\begin{axis}[
grid=major,
width=0.47\columnwidth,
height=0.5\columnwidth,
xlabel=$k$,
ylabel=$\overline{AUC}$,
ymin=0.3, 
ymax=0.9,
xmin=0,
xmax=100,
xticklabels={,0,20,40,60,80,100}]

\addplot[draw=none,color=black,mark=square*] table [x=k,y=LDOF]{Highvs30.dat};\label{pl:LDOF}
\addplot[draw=none,color=red,mark=square]  table [x=k,y=USOS]{Highvs30.dat};\label{pl:USOS}
\addplot[draw=none,color=blue,mark=o] table [x=k,y=tLEF]{Highvs30.dat};\label{pl:tLEF}
\addplot[draw=none,color=blue,mark=square*] table [x=k,y=SOS]{Highvs30.dat};\label{pl:KNNSOS}
\addplot[draw=none,color=red,mark=*] table [x=k,y=ULEF]{Highvs30.dat};\label{pl:ULEF}
\end{axis}
\end{tikzpicture}&
		\begin{tikzpicture}
\footnotesize
\begin{axis}[
grid=major,
width=0.47\columnwidth,
height=0.5\columnwidth,
xlabel=$k$,
ylabel=$\overline{AUC}$,
ymin=0.3, 
ymax=0.9,
xmin=0,
xmax=100,
xticklabels={,0,20,40,60,80,100}]

\addplot[draw=none,color=black,mark=square*] table [x=k,y=LDOF]{Highvs5.dat};\label{pl:LDOF}
\addplot[draw=none,color=red,mark=square]  table [x=k,y=USOS]{Highvs5.dat};\label{pl:USOS}
\addplot[draw=none,color=blue,mark=o] table [x=k,y=tLEF]{Highvs5.dat};\label{pl:tLEF}
\addplot[draw=none,color=blue,mark=square*] table [x=k,y=SOS]{Highvs5.dat};\label{pl:KNNSOS}
\addplot[draw=none,color=red,mark=*] table [x=k,y=ULEF]{Highvs5.dat};\label{pl:ULEF}

\end{axis}
\end{tikzpicture}\\
	\end{tabular}
	\caption{\ref{pl:ULEF} ULEF, \ref{pl:USOS} USOS, \ref{pl:tLEF} tLEF,  \ref{pl:KNNSOS} KNNSOS, \ref{pl:LDOF} LDOF}
	\label{fig:roadResults}
	\vspace{-2mm}
\end{figure}
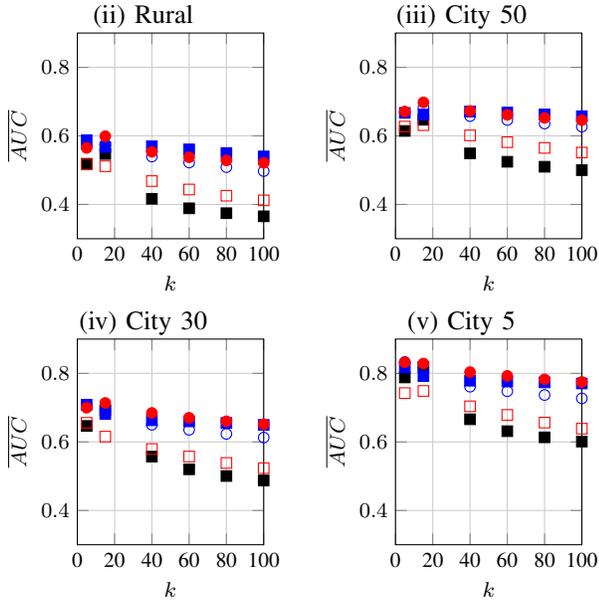

For this application, the number of neighbors is varied as $k\in\lbrace5,15,40,60,80,100\rbrace$. As before, for each $k$, the $AUC$ values of all outliers are averaged as $\overline{AUC}$. The results, when using the classes (ii) to (v) as outliers are depicted in Fig. \ref{fig:roadResults}. The results based on the methods which use the UMAP definitions are colored in red, blue for the Barnes-Hut-SNE and black for the state-of-the-art methods. As it becomes clear from the plots, the scores based on UMAP and Barnes-Hut-SNE are more stable in terms of variations of $k$ than the state-of-the-art scores (Note: Here only LDOF is shown, since it was the overall best performing score in terms of stability). This property is crucial, since in unsupervised tasks, the appropriate number of neighbors is not known and can not be determined as in this setting. Therefore, the following discussion considers the overall performance for all given $k$.

Comparing the plots of the various classes, a clear tendency is that the performance increases from Rural to City 5. This characteristic is intuitive since rural roads are more similar to highway roads than city roads. Overall, the scores based on UMAP and Barnes-Hut-SNE are superior to the state-of-the-art scores. USOS, which shows good performance for the benchmark datasets, performs the worst of the UMAP/Barnes-Hut-SNE based outlier scores in this application. In comparison, the introduced ULEF score performs significantly better. This shows that the ULEF score is preferable to preserve the information about inlier- and outlierness, if UMAP is used. On a comparable level to ULEF performs the Barnes-Hut-SNE based KNNSOS score. To conclude this, based on the application of road infrastructure images one should opt for the novel introduced score ULEF when using UMAP and for KNNSOS when using Barnes-Hut-SNE. The difference in performance between the scores is small.

\begin{figure}
	\vspace{2mm}
	\centering
	\includegraphics[width=0.7\linewidth]{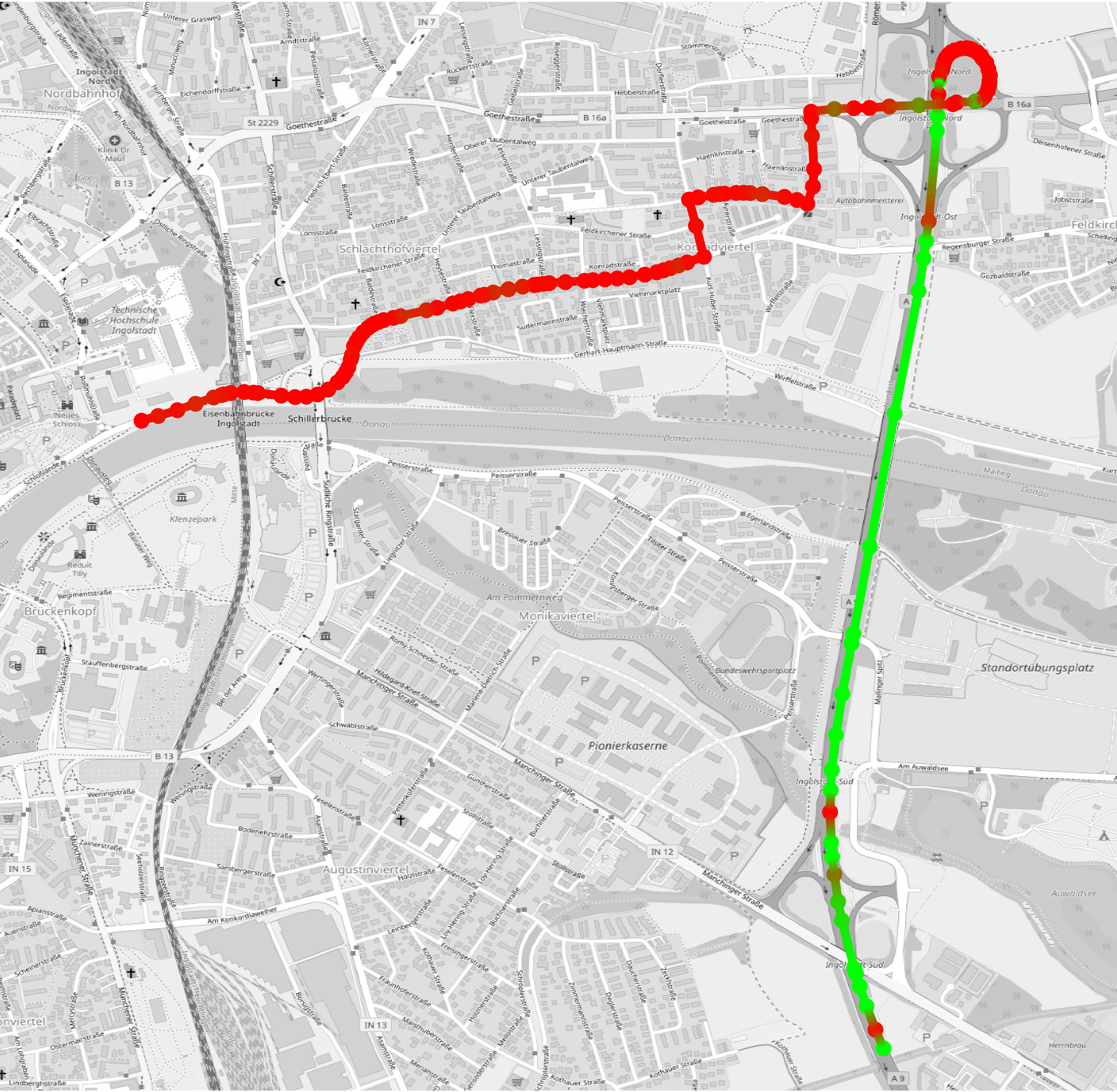}
	\caption{Outlierness for a route. Red: Outlier, Green: Inlier, Base dataset: A9 (highway)}
	\label{fig:route}
	\vspace{-2mm}
\end{figure}

To highlight the capabilities further, a dataset for an exemplary drive is constructed. For this purpose, images for all nodes of the highway A9 from Munich to Ingolstadt-South are collected, but here, the orientation of the images is changed, such that the driving direction is pointing upwards. The driving direction is determined using the position of the current node and the position of the next node on the road. The images collected this way are used as base dataset. Then, using the same orientation correction, the images for the route from Ingolstadt-South to the Technische Hochschule Ingolstadt are extracted and tested with respect to their outlierness. For the first part, this route stays on the highway and then enters inner city infrastructure. In Fig. \ref{fig:route}, the route is depicted, it starts from the bottom. The points depict the used nodes. The color represents the outlierness relative to the range of the outlier scores of the base dataset, where red indicates a high outlier score and green a low. The first part of the route along the highway is dominated from green, and hence already known infrastructure images, which is reasonable, since the base dataset is also a highway (A9). However, some images seem to differentiate even for the highway, since some are marked red. The first red one for example, represents an infrastructure, where an additional lane on the north heading direction is added. Hence, it is most probable that such a constellation is not part of the base dataset. The other red highway points are mainly due to the shape of the merging/leaving lanes. As expected, for the inner city, the most nodes are considered as outliers. However, the first part after the highway tends a bit more towards green. Also this is plausible since this part of the road has two lanes on each side and a separation between the both directions.

\section{Conclusion}\label{sec:conc}
The novel outlier score LEF is presented in this work. The score is designed in such a fashion that it can be embedded into dimensionality reduction techniques which use a directed graph in the first phase. The LEF is applied in UMAP (ULEF) and Barnes-Hut-SNE (tLEF). Besides the LEF, the score USOS is introduced, which is the application of the KNNSOS \cite{Schubert2017a} within UMAP. Both scores strongly focus on the local out- and inlierness. A key factor of both scores is that the same definitions of similarity as in the dimensionality reduction are used. Furthermore, if the dimensionality reduction is performed, the calculation of the scores is of low costs because the required graph has already been constructed. Nevertheless, the scores can be used as standalone, by just constructing the directed graphs without performing the actual embedding.

The scores are applied to benchmark datasets and evaluated in comparison to other $k$ nearest neighbor outlier scores. Comparing the best achieved accuracy, given the optimal number of neighbors, the scores ULEF, tLEF and USOS are at a comparable level with the state-of-the-art methods. In terms of robustness, the new scores are superior to the state-of-the-art scores, since the average accuracy over all $k$s is higher. This fact is of special interest, since in unsupervised tasks, the optimal number of neighbors is not known.

Besides the application to the benchmark dataset, a special focus of this work is the application to novelty detection for road infrastructure images. This is required for identifying new and relevant scenarios, as it is crucial for the validation process of autonomous vehicles. The novel score ULEF alongside with the score KNNSOS are the overall best performing in identifying novel infrastructures. UMAPs advantages \cite{McInnes2018a} in handling large-scale datasets and preserving global data structures, makes ULEF the recommendation of the analyzed methods for the purpose of outlier detection in combination with dimensionality reduction.
\section{Acknowledgments}
Map data copyrighted OpenStreetMap contributors and available from \url{https://www.openstreetmap.org}.
\bibliographystyle{IEEEtran}
\bibliography{ref}

\begin{thebibliography}{10}
\providecommand{\url}[1]{#1}
\csname url@samestyle\endcsname
\providecommand{\newblock}{\relax}
\providecommand{\bibinfo}[2]{#2}
\providecommand{\BIBentrySTDinterwordspacing}{\spaceskip=0pt\relax}
\providecommand{\BIBentryALTinterwordstretchfactor}{4}
\providecommand{\BIBentryALTinterwordspacing}{\spaceskip=\fontdimen2\font plus
\BIBentryALTinterwordstretchfactor\fontdimen3\font minus
  \fontdimen4\font\relax}
\providecommand{\BIBforeignlanguage}[2]{{%
\expandafter\ifx\csname l@#1\endcsname\relax
\typeout{** WARNING: IEEEtran.bst: No hyphenation pattern has been}%
\typeout{** loaded for the language `#1'. Using the pattern for}%
\typeout{** the default language instead.}%
\else
\language=\csname l@#1\endcsname
\fi
#2}}
\providecommand{\BIBdecl}{\relax}
\BIBdecl

\bibitem{Fraade-Blanar2018a}
\BIBentryALTinterwordspacing
L.~Fraade-Blanar, M.~S. Blumenthal, and J.~M. Anderson, \emph{Measuring
  Automated Vehicle Safety: Forging a Framework}.\hskip 1em plus 0.5em minus
  0.4em\relax RAND CORP, 2018. [Online]. Available:
  \url{https://www.ebook.de/de/product/34770704/laura_fraade_blanar_marjory_s_blumenthal_james_m_anderson_measuring_automated_vehicle_safety_forging_a_framework.html}
\BIBentrySTDinterwordspacing

\bibitem{Kalra2016a}
\BIBentryALTinterwordspacing
N.~Kalra and S.~M. Paddock, ``Driving to safety: How many miles of driving
  would it take to demonstrate autonomous vehicle reliability?''
  \emph{Transportation Research Part A: Policy and Practice}, vol.~94, pp. 182
  -- 193, 2016. [Online]. Available:
  \url{http://www.sciencedirect.com/science/article/pii/S0965856416302129}
\BIBentrySTDinterwordspacing

\bibitem{Junietz2018a}
P.~Junietz, W.~Wachenfeld, K.~Klonecki, and H.~Winner, ``Evaluation of
  different approaches to address safety validation of automated driving,'' in
  \emph{2018 21st International Conference on Intelligent Transportation
  Systems (ITSC)}, Nov 2018, pp. 491--496.

\bibitem{Nadarajan2017a}
P.~Nadarajan, M.~Botsch, and S.~Sardina, ``Predicted-occupancy grids for
  vehicle safety applications based on autoencoders and the random forest
  algorithm,'' in \emph{2017 International Joint Conference on Neural Networks
  ({IJCNN})}.\hskip 1em plus 0.5em minus 0.4em\relax {IEEE}, may 2017.

\bibitem{Chaulwar2017b}
A.~Chaulwar, M.~Botsch, and W.~Utschick, ``A machine learning based
  biased-sampling approach for planning safe trajectories in complex, dynamic
  traffic-scenarios,'' in \emph{2017 {IEEE} Intelligent Vehicles Symposium
  ({IV})}.\hskip 1em plus 0.5em minus 0.4em\relax {IEEE}, jun 2017.

\bibitem{Maaten2008a}
L.~v.~d. Maaten and G.~Hinton, ``Visualizing data using t-sne,'' \emph{Journal
  of machine learning research}, vol.~9, no. Nov, pp. 2579--2605, 2008.

\bibitem{JMLR:v15:vandermaaten14a}
\BIBentryALTinterwordspacing
L.~van~der Maaten, ``Accelerating t-sne using tree-based algorithms,''
  \emph{Journal of Machine Learning Research}, vol.~15, pp. 3221--3245, 2014.
  [Online]. Available: \url{http://jmlr.org/papers/v15/vandermaaten14a.html}
\BIBentrySTDinterwordspacing

\bibitem{Tang2016a}
J.~Tang, J.~Liu, M.~Zhang, and Q.~Mei, ``Visualizing large-scale and
  high-dimensional data,'' in \emph{Proceedings of the 25th International
  Conference on World Wide Web - {WWW} 16}.\hskip 1em plus 0.5em minus
  0.4em\relax {ACM} Press, 2016.

\bibitem{McInnes2018a}
L.~{McInnes}, J.~{Healy}, and J.~{Melville}, ``{UMAP: Uniform Manifold
  Approximation and Projection for Dimension Reduction},'' \emph{ArXiv
  e-prints}, Feb. 2018.

\bibitem{Janssens2012a}
J.~Janssens, F.~Huszár, E.~Postma, and J.~van~den Herik, ``Stochastic outlier
  selection,'' Tilburg centre for Creative Computing, techreport 2012-001, Feb.
  2012.

\bibitem{Schubert2017a}
E.~Schubert and M.~Gertz, ``Intrinsic t-stochastic neighbor embedding for
  visualization and outlier detection,'' in \emph{Similarity Search and
  Applications}.\hskip 1em plus 0.5em minus 0.4em\relax Springer International
  Publishing, 2017, pp. 188--203.

\bibitem{Dupuis2015a}
M.~Dupuis \emph{et~al.}, \emph{OpenDRIVE -- Format Specification Rev1.4},
  1st~ed., VIRES Simulationstechnologie GmbH, 2015.

\bibitem{Campos2016a}
G.~O. Campos, A.~Zimek, J.~Sander, R.~J. G.~B. Campello, B.~Micenkov{\'{a}},
  E.~Schubert, I.~Assent, and M.~E. Houle, ``On the evaluation of unsupervised
  outlier detection: measures, datasets, and an empirical study,'' \emph{Data
  Mining and Knowledge Discovery}, vol.~30, no.~4, pp. 891--927, jan 2016.

\bibitem{Ramaswamy2000a}
S.~Ramaswamy, R.~Rastogi, and K.~Shim, ``Efficient algorithms for mining
  outliers from large data sets,'' \emph{{ACM} {SIGMOD} Record}, vol.~29,
  no.~2, pp. 427--438, jun 2000.

\bibitem{Angiulli2002a}
F.~Angiulli and C.~Pizzuti, ``Fast outlier detection in high dimensional
  spaces,'' in \emph{Principles of Data Mining and Knowledge Discovery}.\hskip
  1em plus 0.5em minus 0.4em\relax Springer Berlin Heidelberg, 2002, pp.
  15--27.

\bibitem{Hautamaki2004a}
V.~Hautamaki, I.~Karkkainen, and P.~Franti, ``Outlier detection using k-nearest
  neighbour graph,'' in \emph{Proceedings of the 17th International Conference
  on Pattern Recognition, 2004. {ICPR} 2004.}\hskip 1em plus 0.5em minus
  0.4em\relax {IEEE}, 2004.

\bibitem{Breunig2000a}
M.~M. Breunig, H.-P. Kriegel, R.~T. Ng, and J.~Sander, ``{LOF}: identifying
  density-based local outliers,'' in \emph{Proceedings of the 2000 {ACM}
  {SIGMOD} international conference on Management of data - {SIGMOD} 00}.\hskip
  1em plus 0.5em minus 0.4em\relax {ACM} Press, 2000.

\bibitem{Tang2002a}
J.~Tang, Z.~Chen, A.~W. chee Fu, and D.~W. Cheung, ``Enhancing effectiveness of
  outlier detections for low density patterns,'' in \emph{Advances in Knowledge
  Discovery and Data Mining}.\hskip 1em plus 0.5em minus 0.4em\relax Springer
  Berlin Heidelberg, 2002, pp. 535--548.

\bibitem{Kriegel2009a}
H.-P. Kriegel, P.~Kröger, E.~Schubert, and A.~Zimek, ``{LoOP}: local outlier
  probabilities,'' in \emph{Proceeding of the 18th {ACM} conference on
  Information and knowledge management - {CIKM} 09}.\hskip 1em plus 0.5em minus
  0.4em\relax {ACM} Press, 2009.

\bibitem{Latecki2007a}
L.~J. Latecki, A.~Lazarevic, and D.~Pokrajac, ``Outlier detection with kernel
  density functions,'' in \emph{Machine Learning and Data Mining in Pattern
  Recognition}.\hskip 1em plus 0.5em minus 0.4em\relax Springer Berlin
  Heidelberg, 2007, pp. 61--75.

\bibitem{Schubert2014a}
E.~Schubert, A.~Zimek, and H.-P. Kriegel, ``Generalized outlier detection with
  flexible kernel density estimates,'' in \emph{Proceedings of the 2014 {SIAM}
  International Conference on Data Mining}.\hskip 1em plus 0.5em minus
  0.4em\relax Society for Industrial and Applied Mathematics, apr 2014.

\bibitem{Schubert2012a}
------, ``Local outlier detection reconsidered: a generalized view on locality
  with applications to spatial, video, and network outlier detection,''
  \emph{Data Mining and Knowledge Discovery}, vol.~28, no.~1, pp. 190--237, dec
  2012.

\bibitem{Zhang2009a}
K.~Zhang, M.~Hutter, and H.~Jin, ``A new local distance-based outlier detection
  approach for scattered real-world data,'' in \emph{Advances in Knowledge
  Discovery and Data Mining}.\hskip 1em plus 0.5em minus 0.4em\relax Springer
  Berlin Heidelberg, 2009, pp. 813--822.

\bibitem{Kriegel2008a}
H.-P. Kriegel, M.~S. hubert, and A.~Zimek, ``Angle-based outlier detection in
  high-dimensional data,'' in \emph{Proceeding of the 14th {ACM} {SIGKDD}
  international conference on Knowledge discovery and data mining - {KDD}
  08}.\hskip 1em plus 0.5em minus 0.4em\relax {ACM} Press, 2008.

\bibitem{Houle2018a}
M.~E. Houle, E.~Schubert, and A.~Zimek, ``On the correlation between local
  intrinsic dimensionality and outlierness,'' in \emph{Similarity Search and
  Applications}.\hskip 1em plus 0.5em minus 0.4em\relax Springer International
  Publishing, 2018, pp. 177--191.

\bibitem{Langner2018a}
J.~Langner, J.~Bach, L.~Ries, S.~Otten, M.~Holzapfel, and E.~Sax, ``Estimating
  the uniqueness of test scenarios derived from recorded
  real-world-driving-data using autoencoders,'' in \emph{2018 {IEEE}
  Intelligent Vehicles Symposium ({IV})}.\hskip 1em plus 0.5em minus
  0.4em\relax {IEEE}, jun 2018.

\bibitem{Djenouri2018c}
Y.~Djenouri, A.~Zimek, and M.~Chiarandini, ``Outlier detection in urban traffic
  flow distributions,'' in \emph{2018 {IEEE} International Conference on Data
  Mining ({ICDM})}.\hskip 1em plus 0.5em minus 0.4em\relax {IEEE}, nov 2018.

\bibitem{Djenouri2019a}
Y.~Djenouri, A.~Belhadi, J.~C.-W. Lin, and A.~Cano, ``Adapted k-nearest
  neighbors for detecting anomalies on spatio{\textendash}temporal traffic
  flow,'' \emph{{IEEE} Access}, vol.~7, pp. 10\,015--10\,027, 2019.

\bibitem{McInnes2018b}
L.~McInnes, J.~Healy, N.~Saul, and L.~Grossberger, ``Umap: Uniform manifold
  approximation and projection,'' \emph{The Journal of Open Source Software},
  vol.~3, no.~29, p. 861, 2018.

\bibitem{Janssens2012b}
J.~Janssens, ``scikit-sos,''
  \url{https://github.com/jeroenjanssens/scikit-sos}, 2012.

\bibitem{Campos2016b}
G.~O. Campos, A.~Zimek, J.~Sander, R.~J. G.~B. Campello, B.~Micenková,
  E.~Schubert, I.~Assent, and M.~E. Houle, ``{Supplementary Material for On the
  Evaluation of Unsupervised Outlier Detection: Measures, Datasets, and an
  Empirical Study},''
  {\url{http://www.dbs.ifi.lmu.de/research/outlier-evaluation/DAMI/}},, 2016,
  accessed: 2019-06-27.

\bibitem{OpenStreetMap}
{OpenStreetMap contributors}, ``{Data from 25th october 2019 via Overpass
  API},'' \url{ https://www.openstreetmap.org }, 2019.

\bibitem{SUMO2018}
\BIBentryALTinterwordspacing
P.~A. Lopez, M.~Behrisch, L.~Bieker-Walz, J.~Erdmann, Y.-P. Fl{\"o}tter{\"o}d,
  R.~Hilbrich, L.~L{\"u}cken, J.~Rummel, P.~Wagner, and E.~Wie{\ss}ner,
  ``Microscopic traffic simulation using sumo,'' in \emph{The 21st IEEE
  International Conference on Intelligent Transportation Systems}.\hskip 1em
  plus 0.5em minus 0.4em\relax IEEE, 2018. [Online]. Available:
  \url{https://elib.dlr.de/124092/}
\BIBentrySTDinterwordspacing

\end{thebibliography}
\end{document}